\documentclass[a4paper,fleqn]{cas-sc}

\usepackage[numbers]{natbib}
\usepackage{svg}
\usepackage{booktabs}
\usepackage{caption}
\usepackage{subcaption}
\usepackage{placeins}
\usepackage{float}
\usepackage{url}
\usepackage{fancyref}
\usepackage{hyperref}

\def\tsc#1{\csdef{#1}{\textsc{\lowercase{#1}}\xspace}}
\tsc{WGM}
\tsc{QE}
\tsc{EP}
\tsc{PMS}
\tsc{BEC}
\tsc{DE}

\usepackage{acronym}

\acrodef{NLP}[NLP]{Natural Language Processing}
\acrodef{ML}[ML]{Machine Learning}
\acrodef{SIMON}[SIMON]{SIMilarity-based sentiment projectiON}

\begin{document}

\setcounter{secnumdepth}{4}

\let\WriteBookmarks\relax
\def\floatpagepagefraction{1}
\def\textpagefraction{.001}

\shorttitle{ }

\shortauthors{Andrea Laguna et~al.}


\title[mode = title]{A Cost-aware Study of Depression Language on Social Media using Topic and Affect Contextualization}                      
\tnotemark[1,2]

\tnotetext[1]{This work has been partially supported by AROMA, a subproject of the MIRATAR project (TED2021-132149B-C4), funded by AEI/10.13039/501100011033/, under European Union's NextGenerationEU/PRTR; and by project AMOR, funded by Spanish Ministry of Economic Affairs and Digital Transformation in collaboration with the European Union's NextGenerationEU.}

\author[1]{Andrea Laguna}[orcid=0009-0009-5490-3532]



\ead{andrea.laguna.liang@alumnos.upm.es}


\affiliation[1]{organization={Intelligent Systems Group, ETSI de Telecomunicación, Universidad Politécnica de Madrid},
    addressline={Avenida Complutense, 30}, 
    city={Madrid},
    postcode={28040}, 
    country={Spain}}

\author[1]{Oscar Araque}[
    orcid=0000-0003-3224-0001
]
\cormark[1]
\ead{o.araque@upm.es}

\cortext[cor1]{Corresponding author}

\begin{abstract}
Depression is a growing issue in society's mental health that affects all areas of life and can even lead to suicide.
Fortunately, prevention programs can be effective in its treatment.
In this context, this work proposes an automatic system for detecting depression on social media based on machine learning and natural language processing methods.
This paper presents the following contributions: (i) an ensemble learning system that combines several types of text representations for depression detection, including recent advances in the field; (ii) a contextualization schema through topic and affective information; (iii) an analysis of models' energy consumption, establishing a trade-off between classification performance and overall computational costs.
To assess the proposed models' effectiveness, a thorough evaluation is performed in two datasets that model depressive text.
Experiments indicate that the proposed contextualization strategies can improve the classification and that approaches that use Transformers can improve the overall F-score by 2\% while augmenting the energy cost a hundred times.
Finally, this work paves the way for future energy-wise systems by considering both the performance classification and the energy consumption.
\end{abstract}



\maketitle


\section{Introduction}\label{Intro}

Depression is a common mental disorder that affects 3.8\% of the global population and has a varying prevalence in different demographic groups.
For example, it is more prevalent in adults older than 60 years and in women than men.
This disease can affect all areas of life and, in severe cases, lead to suicide.
However, there are effective treatments for depression, and prevention programs against depression have shown effectiveness~\cite{WHO_dep}.
Depression is said to be underdiagnosed and under-treated, but early detection and treatment can improve its outcome~\cite{Halfin2007}.

Where mental health used to be taboo, now lies a much greater acceptance of the public discussion of mental disorders, in part due to the rise of social media.
This has motivated a boom in discourse on online platforms regarding the topic, and a great abundance of new data that is generated daily.
In this context, we strive to utilize this data to generate knowledge that could help improve the treatment given to people that suffer from depression. 

The objective of this work is to use \acf{NLP} and \acf{ML} techniques to analyse language associated with depression and generate automatic systems capable of classifying the depression shown in texts.
To do so, this work addresses text captured from popular social media platforms.
The aim is to include different contextual information and study what knowledge can be extracted through it.
We study systems in both Spanish and English.
In this way, this study intends to gain insights into the challenge of automatically assessing depression through natural language.
This would allow future work on using such techniques to perform automatic screening on social media, weighting the prevalence of depression in the population.

The objectives of this paper are summarized in the following research questions:

\textbf{RQ1: Can contextualization through emotion and topic information be used for depression detection? }

Depression is a mood disorder that affects a person's emotions~\cite{NIMH_dep}. For this reason, it is expected that contextual information addressing emotion analysis~\cite{wang2020review} will provide useful insight when applied to depression detection.
Also, due to the nature of social media, where a specific type of discourse takes place, the topics about which users with and without depression post about could be indicative of the presence of the disease. Thus, we consider the contextual information provided both by emotion and topic analysis as potentially being relevant for the detection of depression.

\textbf{RQ2: Can we assess model alternatives by means of a trade-off between detection performance and energy efficiency?}

In the last years, great advances have been made both in the realm of \ac{NLP} as well as Machine Learning. These recent, more complex algorithms, often based on neural networks, typically provide better classification performance~\cite{cambria2022deeplearning}.
However, they tend to have a greater computational cost compared to simpler algorithms.
This work studies whether some applications are susceptible of achieving a trade-off between detection performance and energy efficiency, while still being relevant to the task at hand. In this way, we aim to explore whether it is possible to reduce energy costs for NLP and ML applications, and thus reduce the carbon footprint generated by these technologies.

The paper is organized as follows: After this introduction (Section \ref{Intro}), related work covering the topics of word embeddings, emotion lexicons and Transformer models will be presented in Section \ref{related_work}. Next, the models used in this application will be described in Section \ref{model_des}. In Section \ref{evaluation}, the datasets and resources used are described, as well as the experiment design; and the results are presented. In Section \ref{conclusion}, conclusions are presented, as well as limitations and future work. Finally, in Section \ref{appendix}, the reader will find appended material.

\section{Related work} \label{related_work}

In this section, several related works are presented.
However, to the best of our knowledge, there are no works that address depressive language and the study of contextualization and computational costs. 

\subsection{Detection of depression on social media using \ac{NLP} methods}

A study by \citet{article_personality_social_media} found that the language used in social media could provide valuable information regarding the mental life of the person who wrote it, finding that their method could be used to complement traditional methods. This is based on the notion that language contains relevant information about the psychological status and individual personality. Such a method consisted of building a prediction model for personality traits based on texts extracted from the social media platform Facebook. They found the predictions from this model to be coherent with self-reported personality traits and informant reports. In addition, \citet{munmun_twitter} found that social media contains useful information for the detection of depression by studying the publications on the social media platform Twitter of users who reported having a depression diagnosis.  Also, \citet{de2016discovering} showed that it was possible to discern mental health discourse on social media from discourse pertaining to mental illness, specifically, containing suicide risk and ideation. Finally, \citet{coppersmith-etal-2014-quantifying} found that simple natural language processing methods were capable of providing information about mental health and mental disorders. In particular depression, bipolar disorder, post-traumatic stress disorder, and seasonal affective disorder were analysed based on texts extracted from the social media platform Twitter.

In light of these works, it can be seen that many efforts to use social media platforms as tools to detect depression and other mental illnesses have been made.
In a parallel effort, we strive to automatically assess depression in natural language through the exploitation and combination of different representations and knowledge sources.

\subsection{Language Representation}
When processing natural language one of the main challenges is the representation of texts fed to learning models.
Given the nature of such models, it is necessary to transform texts into numerical representations.
To do so, there are several different approaches to perform this mapping \cite{liu2018neural}. From the simple to the more complex, these vectorizers tackle the task in different ways.
In this paper, we explore the use of two approaches: word embeddings and Transformers.

\subsubsection{Word embeddings}
As mentioned, in order to be able to analyse texts with \ac{ML} algorithms, these must be converted into vectors of numbers. Word embeddings, unlike previous techniques, allow the text to be represented as a continuous vector, as opposed to a sparse vector \cite{9390956}. 

Among other characteristics, these types of word vector representations are capable of capturing both semantic and syntactic regularities~\cite{mikolov2012nips}.
These regularities are captured by the offsets between vectors, similar words tend to have similar vectors. So much so, that a particular relationship between words will have an associated offset. For example, the relationship between the vector representations of pairs of the plural and singular of a noun (apples/apple) is close to that of an unrelated noun (cars/car).  Thus, algebraic operations performed on the word vectors reveal meaning. For example, the operation of subtracting the vector representation for "Man" to "King", and adding the vector for "Woman", will yield a vector most closely related to the vector for "Queen" \cite{mikolov-etal-2013-linguistic}. There are many implementations of word embedding models \cite{mikolov2013efficient}.

Word embedding models have been previously applied to the detection of mental illness, specifically anorexia and depression, by \citet{inproceedings_dep_embed}. Similarly to our own approach, pre-trained word embedding models were also used in several implementations which have word embeddings as an input. As such, GloVe Embeddings were used, as well as a custom-trained fastText embedding model. Similarly,  \citet{dep_score_est} also used word embeddings to tackle the problem of depression on texts extracted from the social media Reddit. In their paper, however, rather than a classification, the severity of depression symptoms was estimated.

\subsubsection{Transformers and topic modelling}

The Transformer architecture \cite{vaswani2017attention} uses an encoder-decoder network architecture based only on layers with multi-headed self-attention mechanisms. This architecture originally showed to be superior to others for translation tasks, and it was also proven that it can be successfully generalized to other tasks. These Transformer models are trained on large datasets (generating the so-called pre-trained models) and can be later fine-tuned for specific tasks, resulting in improved performance in many tasks, including language modelling and sentiment analysis. This attention mechanism allows for a greater contextual understanding, and thus allowing for better predictions \cite{Singh2021}. The aim of attention mechanisms is to find long-term dependencies in phrases \cite{gillioz2020overview}.

\citet{10.1145/3578741.3578817} aimed to detect depressive tendencies on the social media platform Twitter. To do so, they used a Transformer model to process the textual features of their texts. This information was then combined with information regarding the online behaviour and interaction of users. This architecture improved the results of their baseline by 12\%.

For the extraction of topics, we use the \textbf{BERTopic} model \cite{grootendorst2022bertopic}. This model was proposed to uncover themes and narratives in texts in an unsupervised manner. It uses a pre-trained transformer-based language model to generate word embeddings, for their capabilities to represent texts in the vector space in a way that makes it possible to obtain their semantic similarity. After the documents have been transformed, the dimensionality of the embeddings is reduced to optimise the clustering that is done afterwards. Finally, using a custom class-based variation of TF-IDF, topic representations are extracted from the clusters of documents. In our paper, this model will be used to extract contextual topic information from the texts. This model has been previously applied in the area of mental health by \citet{Baird2022}. In this paper, BERTopic was applied to texts extracted from Twitter about telehealth for either mental health or substance abuse. 

In addition, BERTopic has also been applied by \citet{arab_twitter} to detect cognitive distortions in the Arabic content of Twitter. This implementation used the previously mentioned word embeddings (implemented by word2vec~\cite{mikolov2013efficient}) together with the topic distribution generated by BERTopic, to generate a contextual topic embedding (CTE). This CTE aimed to both keep the semantic information and the contextual topic representation by concatenating the vectors produced by both of them. The classification performance of the CTE was bench-marked against just a word2vec embedding and was found to improve the results.  

Another work, by \citet{Sarkar2022}, aimed to predict depression and anxiety on the social media platform Reddit with a multi-task learning approach. A combination of word embedding features, using a pre-trained BERT model, and topic modelling features (LDA and BERtopic) were used. This work concluded that this combination of features can be leveraged for domain-specific tasks. 

\subsection{Emotion detection and emotion lexicons}

Emotion Detection or Sentiment Analysis is a field of study that covers emotion detection and recognition from text~\cite{10.1162/COLI_r_00259}.
In this field, we stress two types of analyses. If positive, negative, or neutral feelings are studied, this consists of Sentiment analysis, which aims to determine polarity. Emotion analysis, on the other hand, can detect types of feelings such as happiness, sadness, etc. Emotion models may be dimensional(that is, representing emotion based on its valence, arousal or power) or categorical (where emotions are discrete, such as anger or happiness) \cite{inproceedings_rev_em, Nandwani2021}.

Word-affect association lexicons, also known as emotion lexicons, are a compilation of words associated with the affect that they convey, which includes emotions, sentiments and similar affect concepts. These emotion lexicons may be generated through manual annotation or automatically~\cite{depechemoodpp}. Emotion lexicons have many applications, among which the study of health disorders is most relevant for the case at hand. Lexicon-based emotion analyses have important advantages, such as being interpretable and having a low carbon footprint, which makes it a popular technique for real-world applications \cite{Mohammad2022}. Previously, \citet{Li2022} have generated and used a domain-specific emotion lexicon for the detection of depression, obtaining better results than with general-purpose emotion lexicons.

Another study by \citet{munmun_ppd} used texts extracted from the social media platform Facebook in order to implement prediction systems capable of detecting postpartum depression based on the user's emotional and linguistic expression, as well as their activity and interactions. Contrary to the implementation of our paper, though, the emotions considered were only ``positive affect'' and ``negative affect'', which would be considered as sentiment analysis. This work found that mothers from the cohort suffering from postpartum depression experienced higher levels of negative affect, and lower levels of positive affect, compared with their non-depressed counterparts, although in a less statistically significant way than other previous work in the social media platform Twitter. The authors proposed that the nature of the social media platform chosen could have affected the emotional depression of users. One of these works was done by \citet{munmun_twitter}, where the detection of depression on the social media platform Twitter was studied, and again, positive and negative affect were considered, as well as activation and dominance as determined by the ANEW lexicon~\cite{nielsen2011new}.

\section{Generating Contextualized Representations for Depression Detection} \label{model_des}

In the following paragraphs, an overview of the models used in this implementation will be given. The first model that will be described is a feature extractor named \textbf{\acf{SIMON}}~\cite{ARAQUE2019346}, which uses embedding-based textual representations to generate a fixed length vector, proposing an improvement from regular word-embedding applications. This improvement consists of the following: by using a domain-specific lexicon, the model then represents the words from the input text as a projection to the said lexicon. This projection is done through the semantic similarity of words as given by the word embedding model \cite{ARAQUE2019346}. 

In the original application of this model, an opinion lexicon was used to generate the domain-specific lexicon. However, in this application, the lexicon was generated using the frequency with which words appeared in the texts used for training the model, which corresponded to the training split of the datasets, as was done in the method \textit{FreqSelect} \cite{8962050}. 
In this way, the information encoded in the generated representations contains implicit signals with regard to the objective classes.
As such, the SIMON method extracts distributed representations exploiting both a word embedding model and a domain lexicon. In this application, two embedding models were explored, as will be further described in the following sections.

\begin{figure}[!ht]
     \centering
        \includegraphics[width=\textwidth]{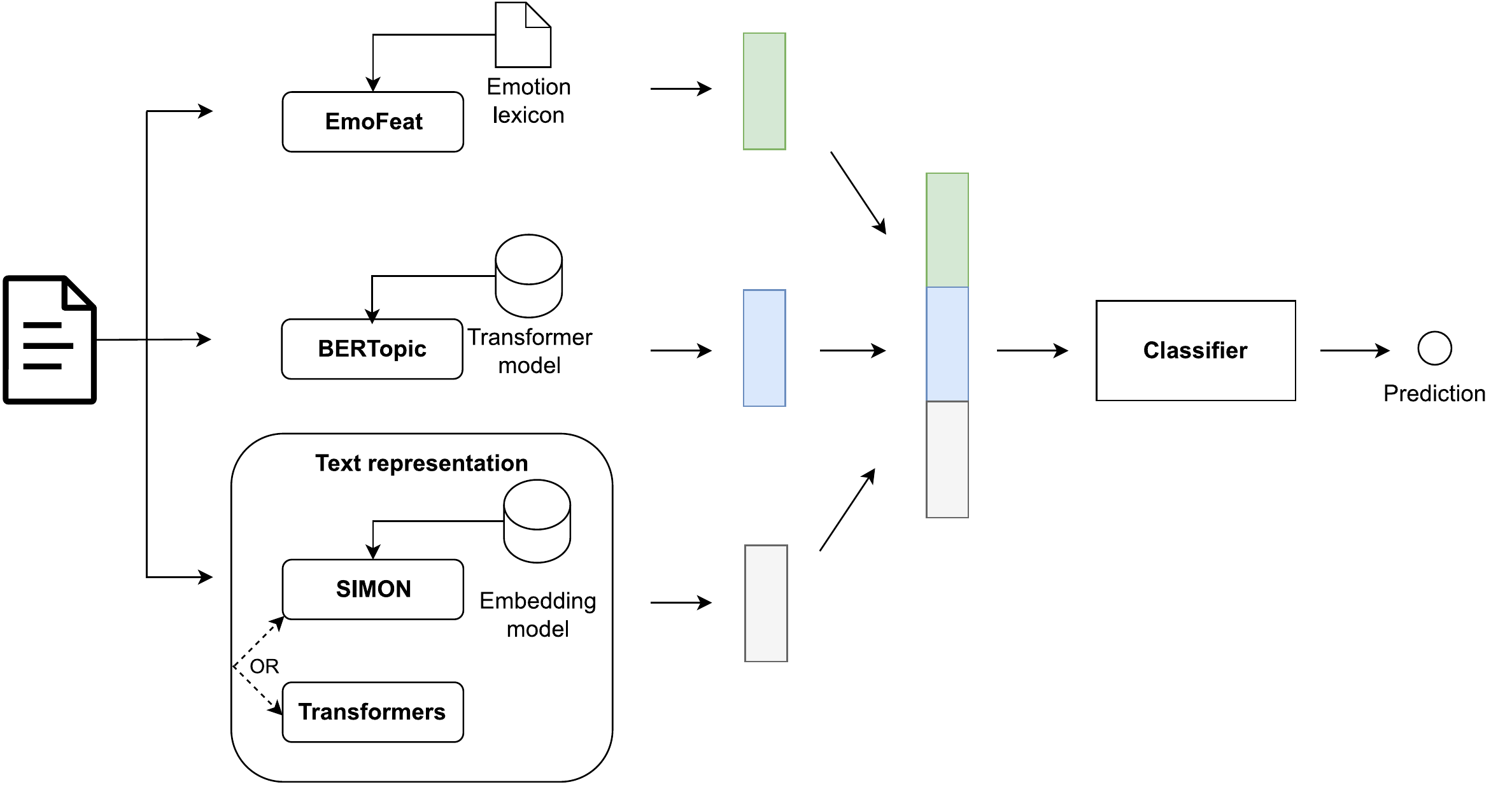}
        \caption{General architecture used in this work.}
        \label{graph:arquict}
        
\end{figure}

\textbf{Transformer Models} \cite{wolf-etal-2020-transformers} were also used in this work to obtain the vector representation for these texts. For the DepSign dataset, the multilingual xlm-roberta-base \cite{DBLP:journals/corr/abs-1911-02116} was used, whilst a Spanish RoBERTa model~\cite{es_model_whatlies} was used for the Spanish dataset~\footnote{\url{https://huggingface.co/PlanTL-GOB-ES/roberta-large-bne}}. 
These transformer models were used without fine-tuning so that overall complexity is reduced in order to generate a vector which could be used as features for a classifier.

Information pertaining to emotions was extracted using \textbf{EmoFeat}  \cite{8962050}.  EmoFeat uses an emotion lexicon for each language to perform a statistical summary of the annotated values for the words contained in a text, thus reducing the matrix generated into a vector. In this particular application, both the maximum and mean emotion were calculated for each emotion in every text. As such, the vectors generated for the texts in English had a length of 16 (as they were 8 annotated emotions), while for the Spanish texts, the final length was of 12. The emotion lexicons used are further described in the next section.

In order to extract topic information to include as additional characteristics for classification, \textbf{BERTopic} was used. This is a topic modelling and visualization library based on Hugging Face Transformers and c-TF-IDF~\cite{grootendorst2022bertopic}. It was used to extract topics from both datasets. In the case of the DepSign dataset, originally 291 topics were extracted by BERTopic. The number of topics was then manually reduced to 128 and 64 from the original topic model, and all three of these cases were then considered for evaluation to see how the size of the topic information vector could affect the classification. In the case of the Spanish dataset, only 21 topics were detected, so no reduction was carried out. Fine-tuning of the model was carried out on the training split of the datasets, and the refined model was used to generate the additional topic characteristic information for all texts in the datasets. For the vectorization, unigrams, bigrams and trigrams were considered. The  graph \ref{graph:arquict} is meant to show a simplified structure of the implementation, featuring the models that were just described.

\section{Evaluation} \label{evaluation}

The proposed models for depression detection have been evaluated using the datasets described in Section~\ref{subsec:datasets}, following the evaluation design that we detail in Section~\ref{subsec:design}.
Following, Section~\ref{subsec:results} thoroughly describes the obtained results, which are aimed at gaining insight into the models' reasoning, their performance and energy consumption.

\subsection{Datasets and resources}
\label{subsec:datasets}

\begin{figure}[!ht]
	 \includegraphics[scale=.45]{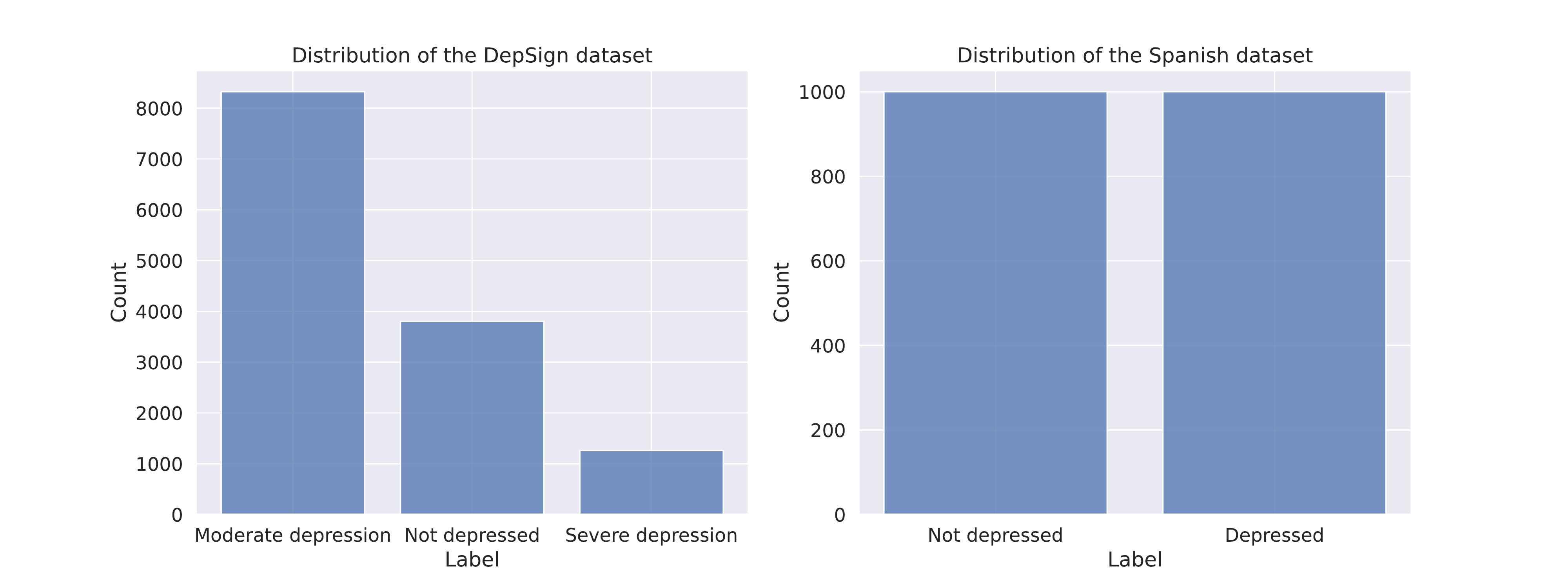}
        \caption{Class distribution of the datasets}
        
    \label{fig:dist}
	
\end{figure}

Two datasets consisting of texts from social media platforms have been selected for this paper. The first dataset is an English dataset made available for the competition ``Detecting Signs of Depression from Social Media Text-LT-EDI@ACL 2022'' \cite{Durairaj2022}, which was created by \citet{10.1007/978-3-031-16364-7_11, s-etal-2022-findings}. In this dataset, the texts were extracted from thematically relevant subreddits from the social media platform Reddit and manually annotated by two domain experts with the following labels: ``not depression'', ``moderate depression'' and ``severe depression''. These classes are unbalanced (see Figure~\ref{fig:dist}), and the total number of texts used was of 13,387. From now on, we refer to this dataset as ``DepSign''.

The second dataset is a Spanish dataset comprised in its entirety of manually selected depressive texts from the social media platform Twitter, provided by \citet{Leis2019}. In order to have both examples of depressive and non-depressive texts, the Spanish dataset was mixed with random tweets in Spanish retrieved from \citet{perez-etal-2022-robertuito}. This dataset was made to be balanced (see Figure~\ref{fig:dist}), and the total amount of instances is 2,000. We denote this dataset as ``Spanish dataset".  A summary of the dataset characteristics can be seen in Table~\ref{table:carac}.

For the application of \ac{SIMON} to the DepSign dataset, the original word2vec~\cite{mikolov2013efficient} embedding model was used.
In contrast, we used the GloVe embeddings from SBWC~\cite{cardellinoSBWCE,pennington-etal-2014-glove} for the Spanish dataset, thus allowing us to represent the Spanish text using embedding similarities.
If we were to represent text in other languages, it is necessary to consider that an embedding model generated for said languages is needed.

In addition to datasets, emotion lexicons were used, both in Spanish and in English. The Spanish lexicon was retrieved from \citet{Rangel2014} and \citet{Sidorov2013}.
This lexicon  has a length of 1,909 words and covers 6 different emotions: anger, disgust, fear, joy, sadness, and surprise.
For English, we use the lexicon from \citet{saifTweets} and \citet{SaifSvetlana}, which has a length of 16,861 tokens and covers 8 emotions: anger, anticipation, disgust, fear, joy, sadness, surprise, and trust.

\begin{table}[!hb]
\begin{minipage}{0.48\textwidth}
    \centering 
    \caption{Characteristics of the datasets}
        \label{table:carac}
        \begin{tabular}{@{}ccc@{}}
        \toprule
                           & \textbf{DepSign dataset} & \textbf{Spanish dataset} \\ \midrule
        \textbf{Number of classes}  & 3               & 2               \\
        \textbf{Number of texts}    & 13387           & 2000            \\
        \textbf{Class distribution} & Unbalanced      & Balanced        \\ \bottomrule
        
        \end{tabular}
      \centering
      
    \end{minipage}
\end{table}

\subsection{Design}\label{subsec:design}
For both datasets, a test-train split approach was followed, with 33\% of the dataset being used for testing. The metric on which the evaluation is assessed is the macro-averaged F1-score, although other metrics, such as the classification report and the F1-score for each class, were also computed and will be used for a finer analysis of the results. Different feature combinations were considered, and their results were compared for both datasets. The feature combinations evaluated are indicated in Table~\ref{tbl1}. The objective of assessing these combinations was to consider the effect these features could have on the classification metrics, as well as other factors, such as the computational costs, which will be described in the results subsection (Section~\ref{subsec:results}).  

Such experimental design raises the question as to why these feature combinations were considered. Firstly, both \ac{SIMON} and Transformers, however differently, extract semantic information from the texts. In addition to this, due to the nature of depression, it made sense to use information on the emotions shown in these texts. Finally, the topics expressed in these texts were considered useful in detecting certain topics of conversation which correlate with depressive symptoms and, as such, could serve as an indication of the presence of the disease, even for a human observer. The reader will find further information on these models in the previous section (Sect.~\ref{model_des}).

\begin{table}[width=.9\linewidth,cols=4,pos=h]
\caption{Combination of features used on the datasets}\label{tbl1}
\begin{tabular*}{\tblwidth}{@{} LLLL@{} }
\toprule
Feature extractor  & Emotions  & Topics\\
\midrule
SIMON & No & No  \\
SIMON & No & Yes  \\
SIMON & Yes & No  \\
SIMON & Yes & Yes \\
Transformers & No & No  \\
Transformers & No & Yes  \\
Transformers & Yes & No  \\
Transformers & Yes & Yes \\

\bottomrule
\end{tabular*}
\end{table}

After the generation of the representations summarised in Table~\ref{tbl1}, these were fed to  different classifiers, and their results were compared.
Each of these scenarios was evaluated with the following \textbf{classifiers}: Forest of randomised trees, KNeighbors Classifier, Linear SVM and Polynomial SVM, as implemented by Sci-Kit learn \cite{Pedregosa2011}. The choice to include several classifiers is based on two main reasons.
Firstly, in order to reduce involuntary bias that may be introduced by the classifiers.
Secondly, to produce a more robust evaluation with more evidence to support it. 

In addition to all the elements mentioned before, we define a \textbf{baseline} as a means for comparison. As such, the baseline consists of a straightforward unigram representation. This benchmark was also evaluated using the four classifiers previously mentioned.

As mentioned before, the computational costs of the generation of each of these features were likewise calculated. This was done with the use of \textbf{CodeCarbon} \cite{codecarbon}. This tool tracks several different metrics for computational costs.
In this work, we use the computation duration and the total energy consumed.
The cost to fine-tune the BERTopic model is considered, but not the costs to train the embedding models or Transformers, which are unknown.
Similarly, we do not consider the associated cost to generate the emotion lexicons used, as we  consider it to be outside the scope of this work.

Finally, in order to obtain a visualisation of the output of the classifiers, the \textbf{SHAP} method \cite{NIPS2017_7062} was used.
This allows us to assess the importance given to different features for a particular classifier, which effectively provides insight into the models' reasoning.

\subsection{Results}\label{subsec:results}

Several things are considered to analyse the results obtained from this set of experiments. Firstly, the value of the macro-averaged F-score is compared for both datasets, as well as all feature combinations. Then, computational costs are considered and weighed in conjunction with F-score values. Finally, the relevance of individual features is analysed.

\subsubsection{Results of the DepSign dataset}

\begin{table}[!ht]
\centering
\resizebox{\textwidth}{!}{%
\begin{tabular}{@{}
>{\columncolor[HTML]{FFFFFF}}l 
>{\columncolor[HTML]{FFFFFF}}c 
>{\columncolor[HTML]{FFFFFF}}l 
>{\columncolor[HTML]{FFFFFF}}c 
>{\columncolor[HTML]{FFFFFF}}l 
>{\columncolor[HTML]{FFFFFF}}l 
>{\columncolor[HTML]{FFFFFF}}l 
>{\columncolor[HTML]{FFFFFF}}l @{}}
\toprule
 &
  \multicolumn{1}{l}{\cellcolor[HTML]{FFFFFF}} &
  \multicolumn{1}{c}{\cellcolor[HTML]{FFFFFF}\textbf{}} &
  \textbf{Dataset} &
  \multicolumn{4}{c}{\cellcolor[HTML]{FFFFFF}\textbf{DepSign}} \\ \midrule
 &
  \multicolumn{1}{l}{\cellcolor[HTML]{FFFFFF}} &
  \multicolumn{1}{c}{\cellcolor[HTML]{FFFFFF}\textbf{}} &
  \textbf{Classifier} &
  \multicolumn{1}{c}{\cellcolor[HTML]{FFFFFF}\textbf{Forest of randomized trees}} &
  \multicolumn{1}{c}{\cellcolor[HTML]{FFFFFF}\textbf{KNeighbors Classifier}} &
  \multicolumn{1}{c}{\cellcolor[HTML]{FFFFFF}\textbf{Linear SVM}} &
  \multicolumn{1}{c}{\cellcolor[HTML]{FFFFFF}\textbf{Polynomial SVM}} \\ \midrule
\multicolumn{1}{c}{\cellcolor[HTML]{FFFFFF}\textbf{Feature extraction}} &
  \textbf{Emotions} &
  \multicolumn{1}{c}{\cellcolor[HTML]{FFFFFF}\textbf{Topics}} &
  \textbf{Number of topics} &
   &
   &
   &
   \\ \cmidrule(r){1-4}
\multicolumn{1}{c}{\cellcolor[HTML]{FFFFFF}\textbf{Unigram baseline}} &
  \textbf{NO} &
  \multicolumn{1}{c}{\cellcolor[HTML]{FFFFFF}\textbf{NO}} &
  \textbf{0} &
  69.39 &
  68.58 &
  51.07 &
  53.80 \\ \cmidrule(r){1-4}
\cellcolor[HTML]{FFFFFF} &
  \cellcolor[HTML]{FFFFFF} &
  \textbf{NO} &
  \textbf{0} &
  70.68 &
  68.11 &
  51.33 &
  59.86 \\ \cmidrule(lr){3-4}
\cellcolor[HTML]{FFFFFF} &
  \cellcolor[HTML]{FFFFFF} &
  \cellcolor[HTML]{FFFFFF} &
  \textbf{64} &
  64.18 &
  \textbf{68.73} &
  53.09 &
  62.04 \\ \cmidrule(lr){4-4}
\cellcolor[HTML]{FFFFFF} &
  \cellcolor[HTML]{FFFFFF} &
  \cellcolor[HTML]{FFFFFF} &
  \textbf{128} &
  62.65 &
  68.61 &
  55.68 &
  62.67 \\ \cmidrule(lr){4-4}
\cellcolor[HTML]{FFFFFF} &
  \multirow{-4}{*}{\cellcolor[HTML]{FFFFFF}\textbf{NO}} &
  \multirow{-3}{*}{\cellcolor[HTML]{FFFFFF}\textbf{YES}} &
  \textbf{All(291)} &
  61.18 &
  68.64 &
  56.91 &
  62.64 \\ \cmidrule(lr){2-4}
\cellcolor[HTML]{FFFFFF} &
  \cellcolor[HTML]{FFFFFF} &
  \textbf{NO} &
  \textbf{0} &
  71.11 &
  68.34 &
  52.16 &
  64.48 \\ \cmidrule(lr){3-4}
\cellcolor[HTML]{FFFFFF} &
  \cellcolor[HTML]{FFFFFF} &
  \cellcolor[HTML]{FFFFFF} &
  \textbf{64} &
  65.89 &
  67.98 &
  54.64 &
  65.51 \\ \cmidrule(lr){4-4}
\cellcolor[HTML]{FFFFFF} &
  \cellcolor[HTML]{FFFFFF} &
  \cellcolor[HTML]{FFFFFF} &
  \textbf{128} &
  64.69 &
  67.66 &
  55.53 &
  65.42 \\ \cmidrule(lr){4-4}
\multirow{-8}{*}{\cellcolor[HTML]{FFFFFF}\textbf{SIMON}} &
  \multirow{-4}{*}{\cellcolor[HTML]{FFFFFF}\textbf{YES}} &
  \multirow{-3}{*}{\cellcolor[HTML]{FFFFFF}\textbf{YES}} &
  \textbf{All(291)} &
  62.33 &
  67.81 &
  57.94 &
  65.63 \\ \cmidrule(r){1-4}
\cellcolor[HTML]{FFFFFF} &
  \cellcolor[HTML]{FFFFFF} &
  \textbf{NO} &
  \textbf{0} &
  73.10 &
  67.96 &
  62.50 &
  68.44 \\ \cmidrule(lr){3-4}
\cellcolor[HTML]{FFFFFF} &
  \cellcolor[HTML]{FFFFFF} &
  \cellcolor[HTML]{FFFFFF} &
  \textbf{64} &
  70.64 &
  67.88 &
  64.92 &
  34.24 \\ \cmidrule(lr){4-4}
\cellcolor[HTML]{FFFFFF} &
  \cellcolor[HTML]{FFFFFF} &
  \cellcolor[HTML]{FFFFFF} &
  \textbf{128} &
  69.16 &
  67.87 &
  58.87 &
  34.24 \\ \cmidrule(lr){4-4}
\cellcolor[HTML]{FFFFFF} &
  \multirow{-4}{*}{\cellcolor[HTML]{FFFFFF}\textbf{NO}} &
  \multirow{-3}{*}{\cellcolor[HTML]{FFFFFF}\textbf{YES}} &
  \textbf{All(291)} &
  68.68 &
  67.81 &
  59.94 &
  34.24 \\ \cmidrule(lr){2-4}
\cellcolor[HTML]{FFFFFF} &
  \cellcolor[HTML]{FFFFFF} &
  \textbf{NO} &
  \textbf{0} &
  \textbf{73.35} &
  67.69 &
  60.47 &
  \textbf{68.48} \\ \cmidrule(lr){3-4}
\cellcolor[HTML]{FFFFFF} &
  \cellcolor[HTML]{FFFFFF} &
  \cellcolor[HTML]{FFFFFF} &
  \textbf{64} &
  70.34 &
  67.61 &
  64.42 &
  34.22 \\ \cmidrule(lr){4-4}
\cellcolor[HTML]{FFFFFF} &
  \cellcolor[HTML]{FFFFFF} &
  \cellcolor[HTML]{FFFFFF} &
  \textbf{128} &
  69.51 &
  67.68 &
  \textbf{65.73} &
  34.22 \\ \cmidrule(lr){4-4}
\multirow{-8}{*}{\cellcolor[HTML]{FFFFFF}\textbf{Transformers}} &
  \multirow{-4}{*}{\cellcolor[HTML]{FFFFFF}\textbf{YES}} &
  \multirow{-3}{*}{\cellcolor[HTML]{FFFFFF}\textbf{YES}} &
  \textbf{All(291)} &
  69.04 &
  67.62 &
  52.53 &
  34.22 \\ \bottomrule
\end{tabular}%
}
\caption{Macro-averaged F-score of the proposed models in the English DepSign dataset.}
\label{tab-EN}
\end{table}

\begin{figure}[!ht]
     \centering
        \includegraphics[scale=.45]{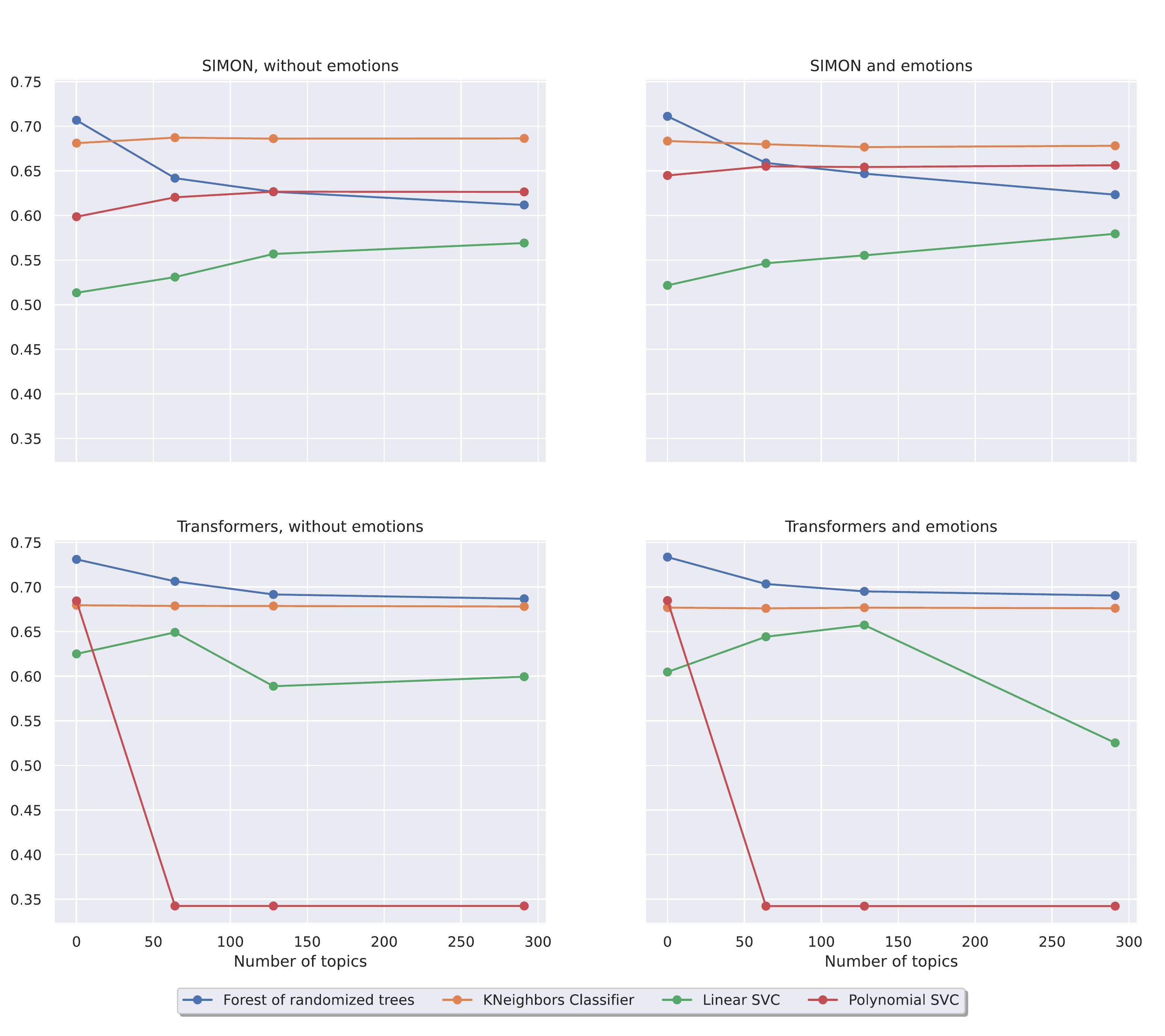}
        \caption{Effect of the number of topics used on the macro-averaged F1-score of the DepSign dataset.}
        \label{fig2}    
\end{figure}

As can be seen in Table~\ref{tab-EN}, the number of topics used for the classification has an effect on its outcome as measured by the \textbf{macro-averaged F-score}. In order to simplify the interpretation of these results, Figure \ref{fig2} shows the effect on such metrics of the number of topics for different combinations of features, for all the classifiers used. The vertical axis shows the macro-averaged F-score, while the horizontal axis shows the number of topics used for that specific classification. Each of the subgraphs shows a different scenario in regard to the features used, as specified in their respective titles. It can be observed that adding topic information can either improve or worsen the classification metric, depending on the classifier used or the number of topics considered.

In regards to the best outcomes according to the macro-averaged F-score, Table \ref{tab-EN} shows that the best results are obtained by the use of Transformers, together with information regarding emotions, but without topics, and the usage of the forest of randomized trees classifier, with a value of \textbf{73.35\%}. However, it is worth mentioning that using this same classifier with \ac{SIMON} and emotion representations nearly achieves the same performance, with an F-score of \textbf{71.11\%}. 

It must be noted that the work that originally presented the DepSign dataset a fundamental analysis was made.
In this analysis, the highest metric was of a weighted average F1-score of 64.7\%~\cite{10.1007/978-3-031-16364-7_11, s-etal-2022-findings}.
In order to compare to such a number, we have also computed the weighted average F1-score which, for our best model, is 79.0\%.

As previously described (see Sect.~\ref{subsec:design}), we thoroughly measure the energy costs of the proposed models.
The normalized results, per processed instance, can be found in Table \ref{tab:costs_EN}.
As seen, the usage of Transformers over \ac{SIMON} has a higher cost both in terms of time and energy of two orders of magnitude. Because the potential applications of the detection of depression on social media could imply the analysis of large amounts of texts, it is worth considering whether this large increase in energy costs is justifiable with a 2\% of improvement in the F-score.

\begin{table}[]
\centering
\resizebox{\textwidth}{!}{%
\begin{tabular}{@{}lllll@{}}
\toprule
{\color[HTML]{000000} \textbf{Model}} &
  {\color[HTML]{000000} \textbf{Normalized training time (s)}} &
  {\color[HTML]{000000} \textbf{Normalized prediction time (s)}} &
  {\color[HTML]{000000} \textbf{Normalized training energy (kW)}} &
  {\color[HTML]{000000} \textbf{Normalized prediction energy (kW)}} \\ \midrule
\rowcolor[HTML]{FFFFFF} 
Transformers & 5.75E-02          & 5.83E-02          & 2.28E-06          & 2.32E-06          \\
\rowcolor[HTML]{FFFFFF} 
SIMON        & 3.98E-04          & 3.98E-04          & 1.49E-08          & 1.48E-08          \\
\rowcolor[HTML]{FFFFFF} 
Emotions     & \textbf{1.27E-04} & \textbf{1.37E-04} & \textbf{3.91E-09} & \textbf{4.21E-09} \\
\rowcolor[HTML]{FFFFFF} 
All topics   & 1.50E-02          & 1.59E-02          & 3.35E-07          & 3.40E-07          \\
\rowcolor[HTML]{FFFFFF} 
128 topics   & 1.89E-03          & 1.09E-02          & 4.05E-08          & 2.35E-07          \\
\rowcolor[HTML]{FFFFFF} 
64 topics    & 1.61E-03          & 1.13E-02          & 3.46E-08          & 2.40E-07          \\ \bottomrule
\end{tabular}%
}

\caption{Computational costs in terms of energy and time consumption for the DepSign dataset. These costs have been normalized, meaning that the values displayed on the table are the time consumed or energy used per text, during the process of either training or prediction, for the different models. }
\label{tab:costs_EN}
\end{table}

\begin{figure}[h!]
     \centering
        \includegraphics[scale=.35]{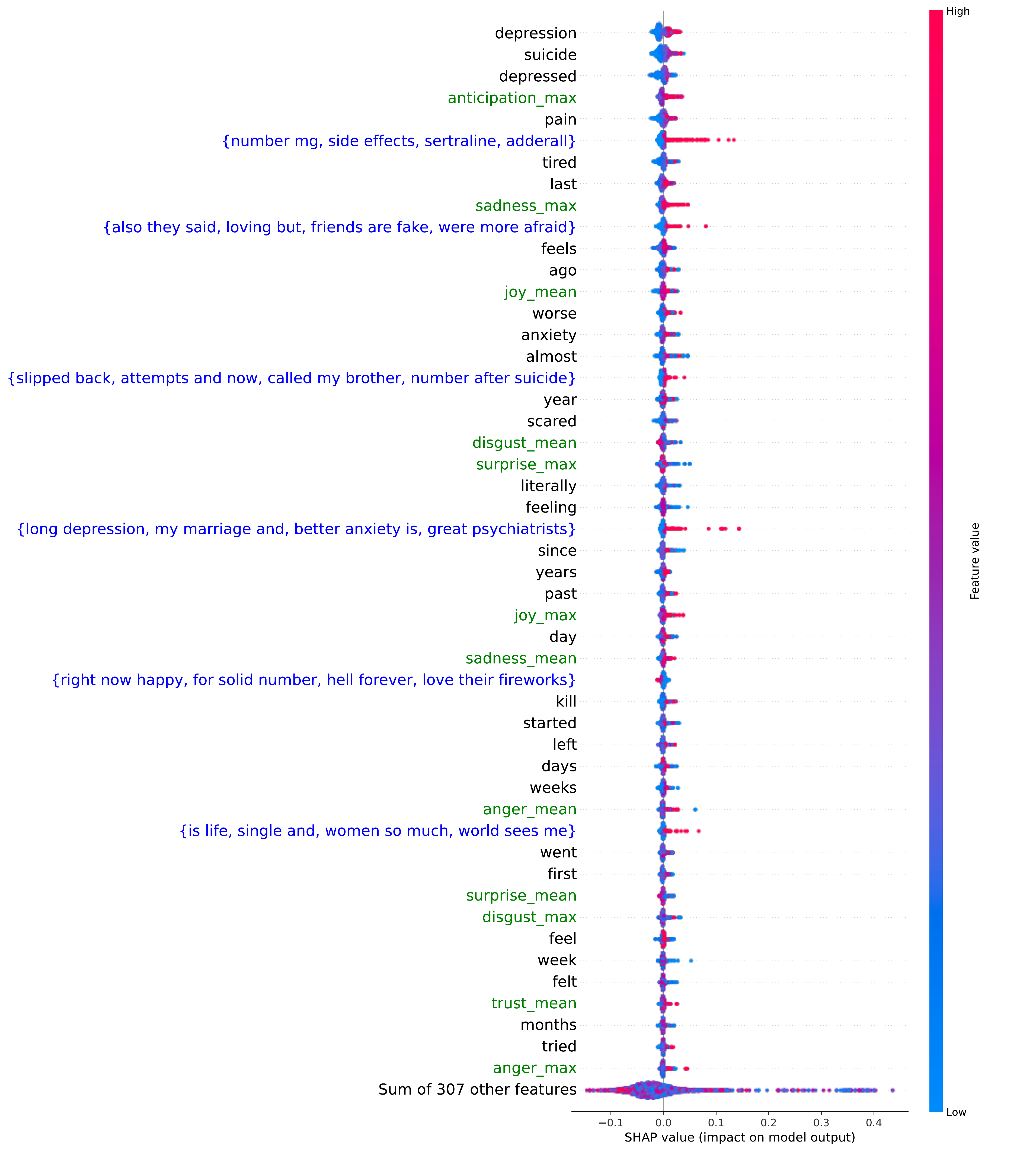}
        \caption{SHAP representation of the Random Forest algorithm application on the DepSign dataset, for the class \textbf{``severe depression''}, considering the features SIMON, emotions and topics. The black tags represent semantic information, while the blue tags represent topic information, and the green ones, emotions.}
        \label{fig_SHAP_EN}     
\end{figure} 

In order to analyze the \textbf{individual feature importance}, Figure \ref{fig_SHAP_EN} was generated using SHAP, as described. It represents the classification criteria for the class ``severe depression'', using the classifier forest of randomized trees. The class ``severe depression'' was chosen, as the diagram produced was deemed to be the most informative. Still, diagrams for the other two classes were likewise generated (see Figure \ref{fig_SHAP_moderate} and Figure \ref{fig_SHAP_not}), which can be found appended in Section \ref{appendix}.  It must be noted that this feature combination is not the best in terms of classification metrics. However, the combination of features it employs is the most informative, which is why it will be used for the following analysis.
The classifier forest of random trees was chosen for this analysis, as has shown to be the option with higher performance scores.
For analyzing the feature importances of the text representation module we select \ac{SIMON} since computing such analyses using larger Transformers models incurs in much higher computational costs.
The interpretation of said analyses is shown in Figure~\ref{fig_SHAP_EN}.

In this figure, the horizontal axis is the SHAP value that encodes the impact on the model output. Thus, all the positive values on the X-axis indicate that that feature, in particular, has an effect on the classification of the text as ``severe depression''. In contrast, the negative values of the x-axis indicate that the text does not belong to the aforementioned class. The colours of the graph, as can be seen on the right border of the image, describe the feature value, with red being high and blue being low, which indicates the intensity of each feature.
Finally, the tags that appear on the left column of the image represent a selection of the most relevant features.
In this way, the texts in blue indicate that that feature belongs to topic information, and the ones in green that the feature represents emotion information. The semantic tokens are represented in  black. The features are listed from more relevant to less relevant for the model. 

Taking this type of representation into account, we consider some of the observations made.
For instance, the blue tag \textit{\{number mg, side effects, sertraline, adderall\}}, which represents a topic regarding medications (sertraline is an antidepressant, while Adderall is used for the treatment of ADHD and narcolepsy), has a very strong effect in favour of classifying this text as ``severe depression''. It is interesting that topic features are so relevant for the classification task, as seen in Figure \ref{fig2}.
Considering the overall performance, we see that the addition of topics may hinder the performance of the model. Despite reducing the macro-averaged F-score, the classifier is using the information from these characteristics in an assiduous manner. 
This indicates that it is not only necessary to add new information sources but model them in a way that the subsequent classifier can efficiently exploit.

By observing Figure~\ref{fig2} it is possible to see which topics indicate the presence of severe depression. These topics include themes such as medications, as well as suicide, fear, anxiety, feeling one´s friends are not genuine, and psychiatrists, among others. These themes are not surprising to find since they correlate well with signs for depression \cite{NIMH_dep}, but having concrete themes which people talk about could be useful to gather signs of alert that could be useful for online monitoring of depression in social media. 

The importance of emotions must also be noted, being some of the most important features. The presence of positive emotions, such as anticipation, could indicate ironic language or noisy instances in the annotation. However, it could also be due to other phenomena, such as mood disorders that can be caused by depression \cite{NIMH_dep}.

To offer some insight into the evaluation, a manual analysis of the texts in the DepSign dataset \cite{10.1007/978-3-031-16364-7_11, s-etal-2022-findings} was performed. The presence of sadness is apparent throughout the whole dataset, but the presence of joy is of more interest for this analysis. Different examples were found.
For example, an instance of a user describing having a good day despite their illness, or a person describing their improvement after beginning therapy.
Other people also described improving with their medication, although sometimes it stopped working after some time. Such an example of a text with words that could have been understood by EmoFeat as having "joy" but, in reality, expresses the opposite, is as follows:

\begin{quote}
    \textit{"Life is overrated imo. : My life is not precious and important. When people say life is precious and important, I just did not understand that and become baffled."}.
\end{quote}

Another case similar to the previous one that also seems to convey some sense of anticipation is:
\begin{quote}
    \textit{"Fantasizing about suicide is the ONLY nice thought I have now. Happy memories, people I've loved, achievements I've had, books I've read and loved, I know they "must have happened", but it's like I can't access that stuff."}    
\end{quote}

In addition, many cases of drug abuse were observed, which would be an example of the aforementioned high-risk activities. 

After this keyword-based manual analysis of the texts, a more thorough computational analysis was made. To do so, the texts were ranked based on their value as given by the different emotion characteristics. Then, the words contained in these texts were compared to the words present in the emotion lexicon, in order to see which words had triggered the emotion. As such, a few examples will be presented, with the trigger words in bold. For instance, for the sadness\_max characteristic, the following text was recovered. It must be noted that four of the trigger words can be seen as semantic information on the SHAP graph.
\begin{quote}
    \textit{``Anyone \textbf{\underline{feel}} their reason/rationality \textbf{\underline{slipping}}?: This is hard to describe, but I feel like the \textbf{\underline{depression}} is overall getting worse because I'm less able to distinguish between ``my voice" and ``the \textbf{\underline{depression}} voice." It wasn't long ago that I would get negative thoughts (my \textbf{\underline{friends}} tolerate me, I'll never reach this goal, etc) and be able to take a second and go ok, this isn't me talking, it's the \textbf{\underline{depression}}. Like I could point out an irrational belief to myself. Now I'm not sure if my thoughts are rational or not, like the ``\textbf{\underline{depressed}} me" and ``me" have merged. I genuinely don't know if it's irrational to think that I'll never be in a relationship, thinking that some \textbf{\underline{suicides}} can be considered rational vs the result of a sick brain, or thinking that my \textbf{\underline{depression}} will only get worse. I thought I would always have reason (barring something like dementia), but even that is questionable at this point and it's scary as hell."}
\end{quote}

The following is also an example where a seemingly positive emotion is detected. In this case, such emotion is joy\_mean, with the trigger words being \textit{happy, goofy, symptoms, life, developing }. Some of these words correspond to what one would expect of positive words, albeit used in a negative context. In contrast, others (\textit{developing, symptoms}) are possibly examples of the previously mentioned noisy instances in the annotation.

\begin{quote}
    \textit{``Desperately need help but can never seem to get results : I’m at a complete loss and have no idea where to look. My whole \textbf{\underline{life}} people have perceived me as the funny, \textbf{\underline{goofy}} extrovert. To be fair it’s a role I’ve given to myself, growing up I would always try to dismiss anything I found negative with a joke. Being the funny one allowed me to hide from reality and kept me \textbf{\underline{happy}}. This worked fine throughout elementary school and made me a pretty \textbf{\underline{happy}} kid, yet in middle school things started taking a turn for the worse. For as I can remember I’ve struggled with ADHD and anger issues. After middle school and throughout high school I started \textbf{\underline{developing}} \textbf{\underline{symptoms}} of anxiety, depression, insomnia, and borderline bipolarism."}
\end{quote}

Similarly, the semantic information has high importance, with the words \textit{depression, suicide} and \textit{depressed} being the three more relevant features. This could indicate that depressed people tend to be quite direct when speaking about their condition online, and thus can indicate what type of online communities should be monitored if one wanted to be able to apply these systems to detect depressed individuals.

Despite being identified as relevant by the classifier, it has been observed that adding contextualization information does not always lead to performance improvement.
This indicates that the proposed contextualization mechanism can be useful for depression detection but needs to be implemented to avoid undesired effects such as overfitting.

\subsubsection{Results of the Spanish dataset}

In the case of the Spanish dataset, the \textbf{macro-averaged F-score} results can be found in Table \ref{tab-ES}. The classifier that obtains the best score is Linear SVM using Transformers without emotion or topic information, with a macro-averaged F-score of \textbf{93.32\%}.
In contrast to the case of the DepSign dataset, the feature extraction done by the Transformer models provides a much higher macro-averaged F-score than the model using \ac{SIMON}, having a difference of roughly 10 points.
While the best score is not achieved using additional context, it can be seen that there are other cases where adding emotion and topic information can lead to performance improvements.
This, again, suggests that these contextualizations can improve overall performance.
It can be observed that for certain combinations of features and classifiers, the unigram baseline can offer a fairly strong performance.

\begin{table}
\centering
\resizebox{\textwidth}{!}{%
\begin{tabular}{@{}
>{\columncolor[HTML]{FFFFFF}}c 
>{\columncolor[HTML]{FFFFFF}}c 
>{\columncolor[HTML]{FFFFFF}}c 
>{\columncolor[HTML]{FFFFFF}}c 
>{\columncolor[HTML]{FFFFFF}}l 
>{\columncolor[HTML]{FFFFFF}}l 
>{\columncolor[HTML]{FFFFFF}}l 
>{\columncolor[HTML]{FFFFFF}}l @{}}
\toprule
\multicolumn{1}{l}{\cellcolor[HTML]{FFFFFF}} &
  \multicolumn{1}{l}{\cellcolor[HTML]{FFFFFF}} &
  \multicolumn{1}{l}{\cellcolor[HTML]{FFFFFF}} &
  \textbf{Dataset} &
  \multicolumn{4}{c}{\cellcolor[HTML]{FFFFFF}\textbf{Spanish}} \\ \midrule
\multicolumn{1}{l}{\cellcolor[HTML]{FFFFFF}} &
  \multicolumn{1}{l}{\cellcolor[HTML]{FFFFFF}} &
  \multicolumn{1}{l}{\cellcolor[HTML]{FFFFFF}} &
  \textbf{Classifier} &
  \multicolumn{1}{c}{\cellcolor[HTML]{FFFFFF}\textbf{Forest of randomized trees}} &
  \multicolumn{1}{c}{\cellcolor[HTML]{FFFFFF}\textbf{KNeighbors Classifier}} &
  \multicolumn{1}{c}{\cellcolor[HTML]{FFFFFF}\textbf{Linear SVM}} &
  \multicolumn{1}{c}{\cellcolor[HTML]{FFFFFF}\textbf{Polynomial SVM}} \\ \midrule
\textbf{Feature extraction} &
  \textbf{Emotions} &
  \textbf{Topics} &
  \textbf{Number of topics} &
   &
   &
   &
   \\ \cmidrule(r){1-4}
\textbf{Unigram baseline} &
  \textbf{NO} &
  \textbf{NO} &
  \textbf{0} &
  83.18 &
  79.59 &
  82.87 &
  80.42 \\ \cmidrule(r){1-4}
\cellcolor[HTML]{FFFFFF} &
  \cellcolor[HTML]{FFFFFF} &
  \textbf{NO} &
  \textbf{0} &
  82.27 &
  77.11 &
  80.75 &
  80.57 \\ \cmidrule(lr){3-4}
\cellcolor[HTML]{FFFFFF} &
  \multirow{-2}{*}{\cellcolor[HTML]{FFFFFF}\textbf{NO}} &
  \textbf{YES} &
  \textbf{All (21)} &
  83.30 &
  79.08 &
  82.42 &
  82.57 \\ \cmidrule(lr){2-4}
\cellcolor[HTML]{FFFFFF} &
  \cellcolor[HTML]{FFFFFF} &
  \textbf{NO} &
  \textbf{0} &
  82.42 &
  75.28 &
  79.84 &
  79.24 \\ \cmidrule(lr){3-4}
\multirow{-4}{*}{\cellcolor[HTML]{FFFFFF}\textbf{SIMON}} &
  \multirow{-2}{*}{\cellcolor[HTML]{FFFFFF}\textbf{YES}} &
  \textbf{YES} &
  \textbf{All (21)} &
  83.31 &
  77.40 &
  82.73 &
  80.61 \\ \cmidrule(r){1-4}
\cellcolor[HTML]{FFFFFF} &
  \cellcolor[HTML]{FFFFFF} &
  \textbf{NO} &
  \textbf{0} &
  90.14 &
  \textbf{89.67} &
  \textbf{93.32} &
  \textbf{83.57} \\ \cmidrule(lr){3-4}
\cellcolor[HTML]{FFFFFF} &
  \multirow{-2}{*}{\cellcolor[HTML]{FFFFFF}\textbf{NO}} &
  \textbf{YES} &
  \textbf{All (21)} &
  \textbf{90.60} &
  88.29 &
  90.76 &
  83.33 \\ \cmidrule(lr){2-4}
\cellcolor[HTML]{FFFFFF} &
  \cellcolor[HTML]{FFFFFF} &
  \textbf{NO} &
  \textbf{0} &
  89.99 &
  89.37 &
  93.02 &
  \textbf{83.57} \\ \cmidrule(lr){3-4}
\multirow{-4}{*}{\cellcolor[HTML]{FFFFFF}\textbf{Transformers}} &
  \multirow{-2}{*}{\cellcolor[HTML]{FFFFFF}\textbf{YES}} &
  \textbf{YES} &
  \textbf{All (21)} &
  90.60 &
  88.45 &
  90.30 &
  83.33 \\ \bottomrule
\end{tabular}%
}
\caption{Macro-averaged F-score of the proposed models in the Spanish dataset.}
\label{tab-ES}
\end{table}

The \textbf{computational costs} for this dataset can be found in Table \ref{tab:cost_ES}. In this case, the normalized training time, training energy and prediction energy are three orders of magnitude larger for the Transformer models. In comparison, the normalized prediction energy is two orders of magnitude greater. It is possible that the amount of time and energy required are not linear during the execution of these models. Because this dataset has a much smaller size, the costs are less distributed among the samples. Again, this increase in costs must be weighed against the improvement of the classification capabilities of the method.

\begin{figure}[h!]
     \centering
        \includegraphics[width=\textwidth]{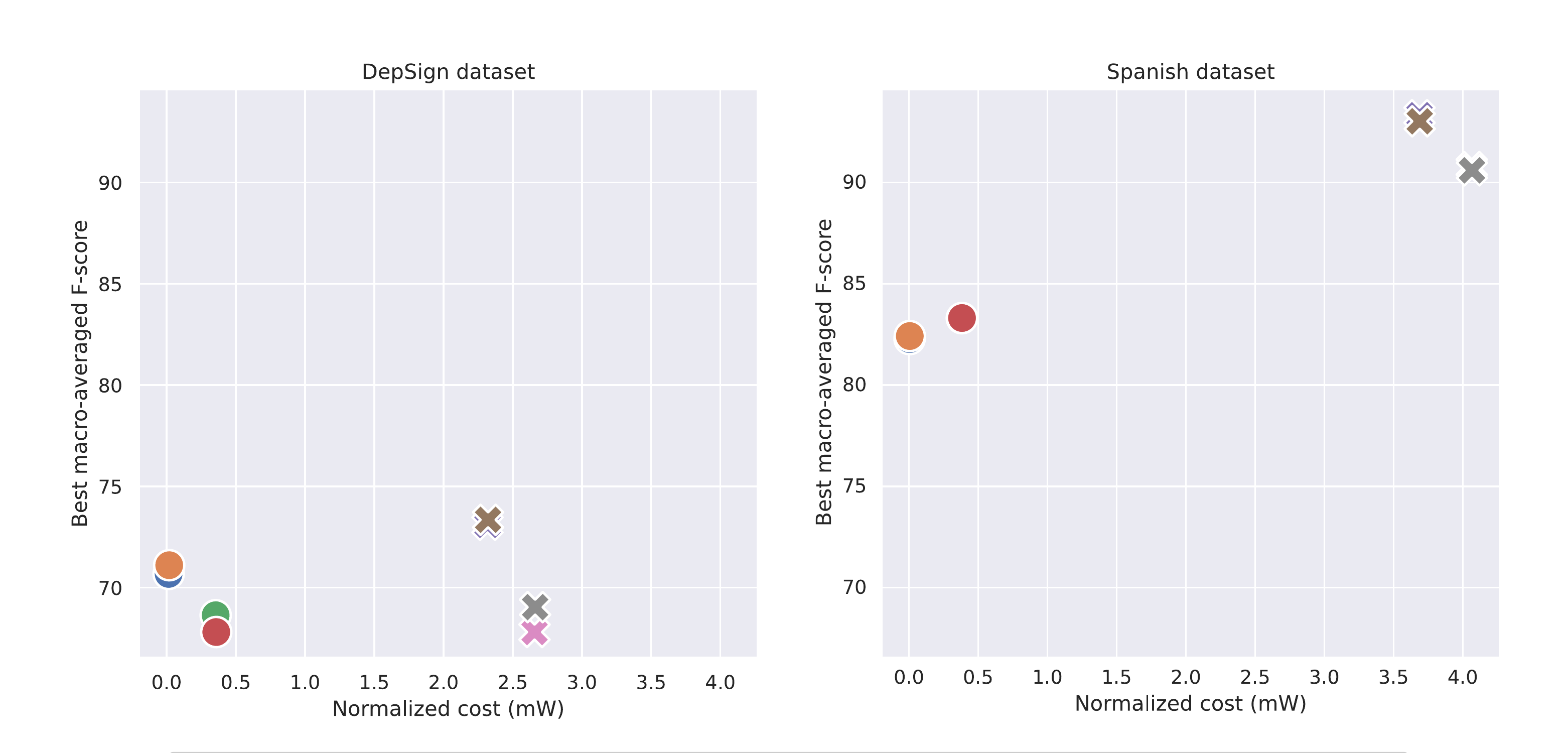}
        \caption{Comparison of the normalized cost in terms of energy consumed per text processed in milliwatts during the prediction phase, plotted against the best F-score obtained by that feature combination.}
        \label{fig:costs_BOTH}  
\end{figure}

In order to summarize the information about costs and performance, Figure~\ref{fig:costs_BOTH} displays the best F-score obtained for any given combination of characteristics against their cost (in mW). This graph summarizes this information for both datasets, displaying the qualities that have been previously discussed. Firstly, it is possible to see that the DepSign dataset achieves a lower F-score than the Spanish dataset. In addition, it can be seen that, for both cases, the usage of Transformer models incurs in a higher cost. In the case of the Spanish dataset, these are greater, possibly for the reasons that have already been mentioned.

Another relevant aspect regarding energy consumption is that, similarly to what we observed in the DepSign dataset, the energy required to extract emotion features is lower than all other methods used.
These results respond to the simplicity of the EmoFeat method.
Also, this indicates that adding such a method with little extra energy cost consistently improves the results obtained in the DepSign dataset. However, this is not always the case for the Spanish dataset since adding emotion information can reduce the performance score. Since the Spanish emotion lexicon is of a much smaller size than the DepSign dataset, we hypothesize that this difference may be caused by the Spanish emotion lexicon's reduced coverage.

\begin{table}[]
\centering
\resizebox{\textwidth}{!}{%
\begin{tabular}{@{}lllll@{}}
\toprule
{\color[HTML]{000000} \textbf{Model}} &
  {\color[HTML]{000000} \textbf{Normalized training time (s)}} &
  {\color[HTML]{000000} \textbf{Normalized prediction time (s)}} &
  {\color[HTML]{000000} \textbf{Normalized training energy (kW)}} &
  {\color[HTML]{000000} \textbf{Normalized prediction energy (kW)}} \\ \midrule
\rowcolor[HTML]{FFFFFF} 
{\color[HTML]{000000} \textbf{Transformers}} &
  {\color[HTML]{000000} 1.02E-01} &
  {\color[HTML]{000000} 9.40E-02} &
  {\color[HTML]{000000} 3.99E-06} &
  {\color[HTML]{000000} 3.69E-06} \\
\rowcolor[HTML]{FFFFFF} 
{\color[HTML]{000000} \textbf{SIMON}} &
  {\color[HTML]{000000} 1.45E-04} &
  {\color[HTML]{000000} 1.40E-04} &
  {\color[HTML]{000000} 4.62E-09} &
  {\color[HTML]{000000} 4.79E-09} \\
\rowcolor[HTML]{FFFFFF} 
{\color[HTML]{000000} \textbf{Emotions}} &
  {\color[HTML]{000000} \textbf{4.69E-05}} &
  {\color[HTML]{000000} \textbf{6.13E-05}} &
  {\color[HTML]{000000} \textbf{1.05E-09}} &
  {\color[HTML]{000000} \textbf{1.56E-09}} \\
\rowcolor[HTML]{FFFFFF} 
{\color[HTML]{000000} \textbf{All topics}} &
  {\color[HTML]{000000} 2.17E-02} &
  {\color[HTML]{000000} 1.79E-02} &
  {\color[HTML]{000000} 4.57E-07} &
  {\color[HTML]{000000} 3.77E-07} \\ \bottomrule
\end{tabular}%
}
\caption{Computational costs for the Spanish dataset.}
\label{tab:cost_ES}
\end{table}

\begin{figure}[h!]
     \centering
        \includegraphics[scale=.35]{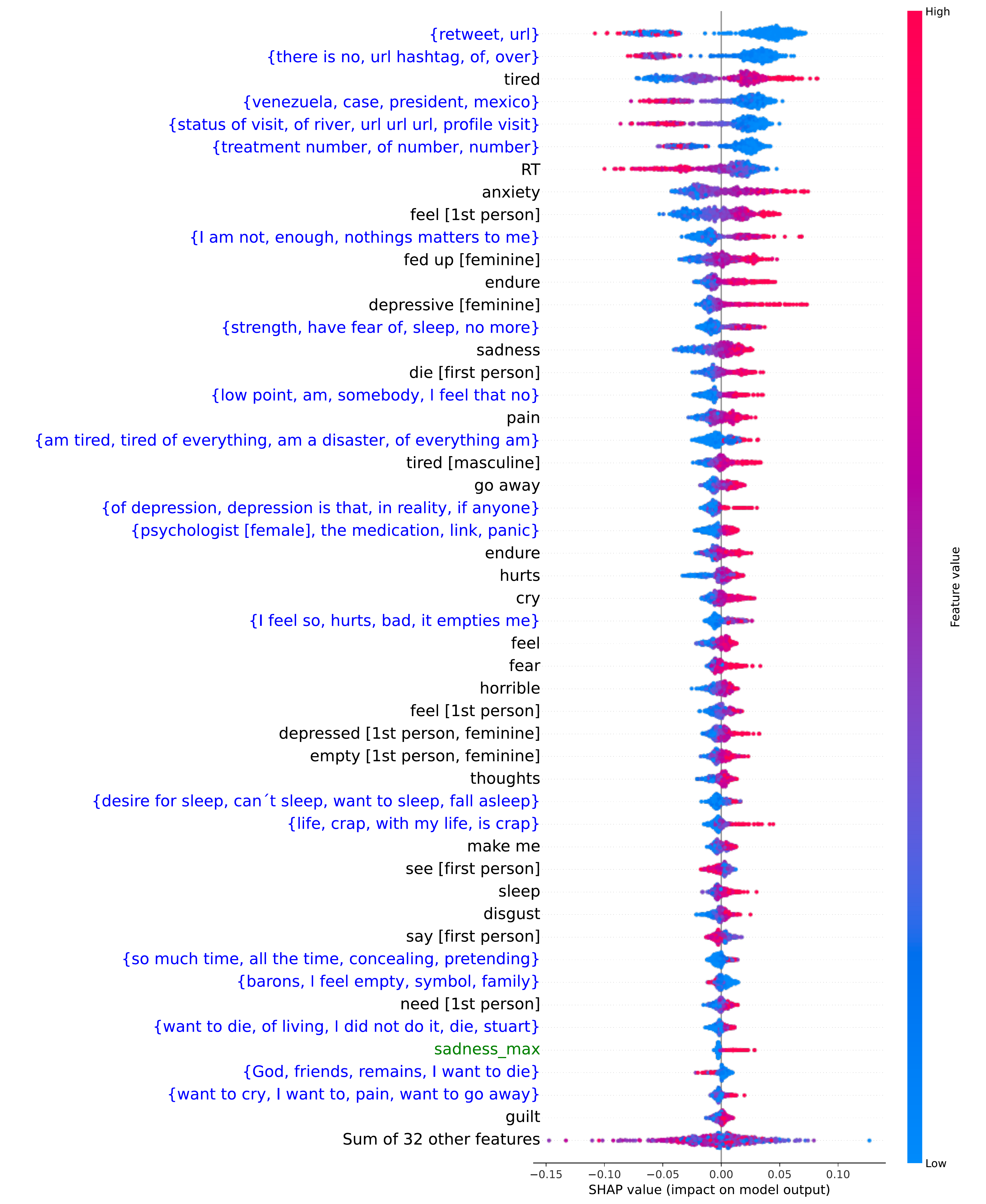}
        \caption{Translated SHAP representation of the Random Forest algorithm application on the \textbf{Spanish dataset}, considering the features SIMON, emotions, and topics. The black tags represent semantic information, the blue tags represent topic information, and the green ones, emotions.}
        \label{fig_SHAP_ES}  
\end{figure}

Following, Figure \ref{fig_SHAP_ES} illustrates the SHAP analysis for \textbf{individual feature importance}~\footnote{As the dataset contains Spanish texts, the labels for this image have been translated to English, trying to preserve the details implied in the Spanish words. As such, where there was no direct word-for-word translation, more words may have been used, or clarification between square brackets included.}.
As before, topics are coloured in blue, emotions are coloured in green, and semantic information is in black. The forest of randomized trees algorithm was also used for this diagram. 
Because this dataset contains a binary classification task, the positive SHAP values can be interpreted as indicating depression, while the negative ones indicate its absence. When analysing this dataset, it can be seen that the emotions display a lower importance for classification, with only \textit{"sadness\_max"} appearing in the tags displayed and indicating the presence of depression. Again, the reduced importance of emotions in the Spanish dataset compared to the DepSign dataset could be due to a decreased coverage of the emotion lexicon (see Sect.~\ref{subsec:datasets}).

About the topic information, the first two items on the graph (\textit{\{retweet, url\}}, \textit{\{there is no, url hashtag, of, over\}}) are topics that convey information about behaviours and language about this specific social media. The presence of these topics indicates that it is likely that the tweet is not depressive, but their absence does not give information about the classification of the text. Still, it provides information on the usage of links to outside content, hashtags, and retweets that could be useful for initial filtering. 

Both topic and semantic information relate to signs of depression, such as problems with sleep, tiredness, and hopelessness \cite{NIMH_dep}.
This suggests that said information is indicative of depression.
While emotion contextual information is used to some degree to classify this dataset, the contextual information provided by topic information is much more relevant.

\section{Conclusions} \label{conclusion}

This work presents a machine learning method for automatically assessing depression in natural language.
As a pivotal component of this method, we consider the contextualization through two relevant information sources: affect and topic analysis.
It has been seen that adding such additional representations can enhance performance, although this depends on the specific learning model used.
Furthermore, we consider the energy consumption of the proposed models, with a focus on finding a trade-off between classification performance and computational complexity, aiming at usability.
We have thoroughly evaluated the proposed models using two datasets.

In light of the analysis of the used data, we draw several conclusions. Answering \textbf{RQ1}, as mentioned during the study of the results for both datasets, contextualization through emotion and topic information has proven to be informative in the detection of depression, being capable of generating valuable information beyond the current task. This information has been shown to relate well to signs that indicate depression but is further capable of providing concrete examples of behaviours that may be indicative of its presence.
Specific procedures on how to include such contextual information need to be developed since the improvements are not always consistent across learners.

Addressing research question \textbf{RQ2}, we have identified  a clear candidate application in our experiments in which the trade-off between computational costs and classification performance is appropriate. Thus, while it is undeniable that state-of-the-art techniques provide a unique opportunity for applications in many fields, it must be noted that, for others, such advanced techniques may not be necessary as other, pre-existing techniques, offer a better balance between computational cost and results for suitable scenarios. 

Finally, it is worth mentioning that the interpretability of classifiers is of great value as the analyses performed in this paper depend on the interpretability of the different learners.
This allows us to gain insights into the classification process and extract aggregated knowledge.
As such, not only can this information be applied for further uses, but the functioning of the algorithm can be better assessed and understood. This is of great importance, specifically in any areas related to health, as the implementation of these methods will require the understanding and acceptance of clinicians.

The are some limitations to this work that must be noted.
Firstly, the datasets used the model ``depression'' as a discrete category, but the depression condition is far more complex, including major depression and persistent depressive disorder, among others \cite{NIMH_dep}.
This nuisance may affect the generalization capabilities of the developed learning models, as it simplifies the original problem. Finally, simple learning algorithms were applied to all models in order to preserve their interpretability and thoroughly study the effect of the computed features, avoiding the side effect of more complex methods.
However, it is likely that more complex learning algorithms would have outperformed the ones chosen.
Instead, this work addresses the study of depressive language, and thus we have not aimed to maximize the classification performance. 

Despite these limitations, we consider that the models described in this paper offer useful insights. For future work, in order to surpass these limitations, the usage of more nuanced datasets that make a distinction between types of depression is proposed. This would require clinical diagnosis by trained professionals. This interdisciplinary approach of cooperation with mental health professionals could improve the outcomes of these experiments, as well as the quality of the knowledge extracted. In addition, in this paper, we have considered depression as a static label. However, this disease develops over a period of time. For this reason, further analysis of users' posting history, that is, taking time variations into account, could be used to detect early signs of the disorder. Furthermore, additional testing combining the text representations provided by SIMON and Transformer models could be used to analyze whether the combination of both methods could provide additional information. 

\clearpage
\bibliographystyle{cas-model2-names}

\appendix
\clearpage
\section{Appendix} \label{appendix}
\subsection{SHAP graph for the "moderate depression" class of the DepSign dataset}

\begin{figure}[h!]
     \centering
        \includegraphics[scale=.35]{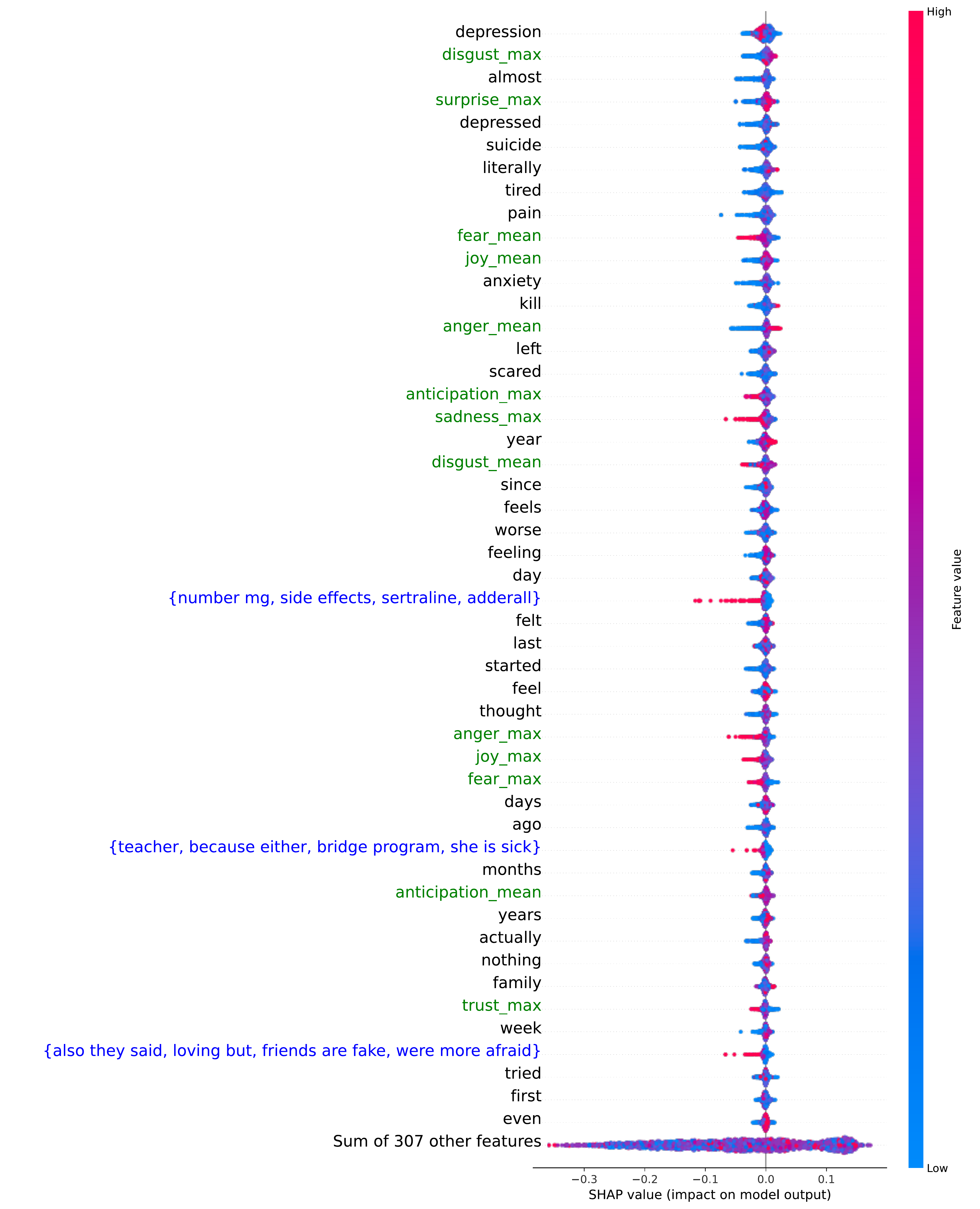}
        \caption{SHAP representation of the Random Forest algorithm application on the DepSign dataset, considering the features SIMON, emotions and topics, for the class \textbf{"moderate depression"}. The black tags represent semantic information, while the blue tags represent topic information, and the green ones, emotions.}
        \label{fig_SHAP_moderate}
        
\end{figure} 

\FloatBarrier
\clearpage
\subsection{SHAP graph for the "not depression" class of the DepSign dataset}

\begin{figure}[h!]
     \centering
        \includegraphics[scale=.35]{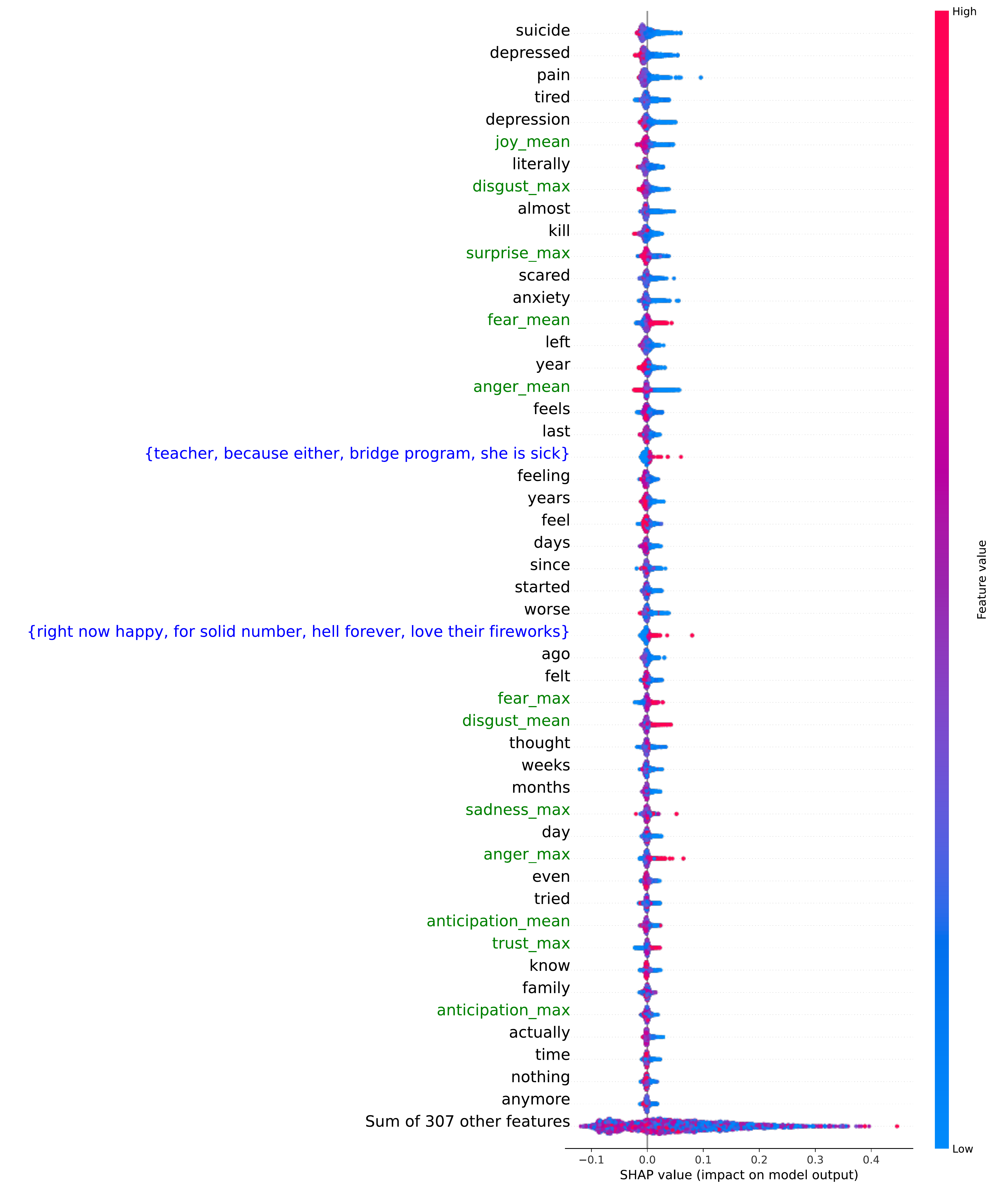}
        \caption{SHAP representation of the Random Forest algorithm application on the DepSign dataset, considering the features SIMON, emotions and topics, for the class \textbf{"not depression"}. The black tags represent semantic information, while the blue tags represent topic information, and the green ones, emotions.}
        \label{fig_SHAP_not}
        
\end{figure} 

\FloatBarrier

\end{document}